\documentclass[10pt,twocolumn]{article}
\usepackage[letterpaper,margin=0.72in,top=0.68in,bottom=0.82in,columnsep=0.24in]{geometry}
\usepackage{times}
\usepackage{cite}
\usepackage{amsmath,amssymb,amsfonts}
\usepackage{graphicx}
\usepackage{array}
\usepackage{multirow}
\usepackage{CJKutf8}
\usepackage{textcomp}
\usepackage{url}
\usepackage{microtype}
\usepackage[font=small,labelfont=bf]{caption}
\usepackage{stfloats}
\usepackage{placeins}
\usepackage{enumitem}
\usepackage{indentfirst}
\newcommand{\setsmocapdisplaymathspacing}{%
  \setlength{\abovedisplayskip}{4pt plus 1pt minus 1pt}%
  \setlength{\belowdisplayskip}{4pt plus 1pt minus 1pt}%
  \setlength{\abovedisplayshortskip}{2pt plus 1pt minus 1pt}%
  \setlength{\belowdisplayshortskip}{4pt plus 1pt minus 1pt}%
  \setlength{\jot}{1.5pt}%
}
\makeatletter
\g@addto@macro\normalsize{\setsmocapdisplaymathspacing}
\makeatother
\def\BibTeX{{\rm B\kern-.05em{\sc i\kern-.025em b}\kern-.08em
    T\kern-.1667em\lower.7ex\hbox{E}\kern-.125emX}}
\title{\vspace{-0.35in}\bfseries SmoCap: Movement Reconstruction under Morphology--Pose Ambiguity through Unified Scale--Pose Canonicalization}
\author{Shihao Li$^{*}$ \quad Naohiko Sugita\\
\small The Research into Artifacts, Center for Engineering, The University of Tokyo\\
\small $^{*}$Corresponding author: lshdlut@gmail.com}
\date{}

\begin{document}
\raggedbottom
\twocolumn[{%
\maketitle
\begin{center}
\begin{minipage}{0.92\textwidth}
\noindent{\large\bfseries Abstract}\par\vspace{0.45em}
\noindent Movement reconstruction pipelines need estimates that support interpretable joint motion and subject morphology, not only low marker fitting error. The same marker fitting error can be explained by different mixtures of morphology and posture, while weakly observed degrees of freedom are not uniquely identifiable, yielding anatomically inconsistent yet numerically acceptable solutions. We present SmoCap, a unified scale--pose framework that resolves morphology--posture ambiguity while preserving coordinated motion under weak observability. SmoCap solves a constrained trust-region QP with analytical proxy-mapped pose and scale Jacobians. The low dimensional proxy map stabilizes weakly observed directions and drives coordinated structures. Under matched observations, SmoCap is compared with an established OpenSim baseline using fluoroscopy-derived knee motion and anthropometric ground truth as external references. Extreme yoga sequences further probe coordinated spine motion under weak observability. In the controlled comparison, the OpenSim baseline achieved lower marker RMSE (8.81 vs 19.14 mm), whereas SmoCap achieved lower knee-orientation RMSE against fluoroscopy (5.82$^\circ$ vs 7.50$^\circ$) in 24 of 28 trials and all six subjects. Mean absolute anthropometric endpoint errors were 8.44 vs 9.51 mm on CAMS-Knee and 6.87 vs 8.77 mm on Riglet for SmoCap and OpenSim, respectively. Proxy coupling preserved expressive and coordinated spine motion with marginal fitting error increase (+0.14 mm, +0.6\%) in the yoga ablation. Median runtime was 0.204--0.332 ms/frame, with consistently 2--3 iterations. Marker RMSE alone did not reliably indicate better motion or morphology recovery. SmoCap instead combines unified scale--pose estimation with proxy coordination. It achieved lower motion and morphology errors in the external evaluations while supporting weak-observability-aware, cohort-scale movement reconstruction for downstream pipelines that depend on interpretable subject motion and morphology.
\par\vspace{0.55em}
\end{minipage}
\end{center}
\vspace{1.0em}
}]
\section{Introduction}

Movement reconstruction pipelines need interpretable estimates of joint motion and subject morphology, yet reconstruction quality is routinely judged by marker fitting error. However, the same marker or keypoint tracking error can be explained by different mixtures of shape and posture change. The resulting trajectories are then anatomically inconsistent: segment lengths drift, joint behavior varies across sequences for the same subject, and the canonical motion representation breaks down. Here, motion canonicalization means mapping motions from different subjects and sessions into a shared and consistent parameter representation. There have been efforts to transform heterogeneous motion capture datasets into unified parameterizations (e.g., SMPL-based archives such as AMASS), highlighting the value of motion canonicalization \cite{loper2015smpl,mahmood2019amass}. The same need extends to datasets that unify movement and ground-reaction data \cite{han2023groundlink}. At the same time, biomechanics and human body reconstruction increasingly require larger numbers of canonicalized human motions with coherent morphology and kinematics in a unified framework \cite{johnson2019workload,fregly2007gaitmod}. This highlights the need for reliable scale--pose canonicalization at scale, with high throughput, practical accessibility, and validity beyond marker fit.

\begin{figure*}[!t]
\centering
\includegraphics[width=\linewidth]{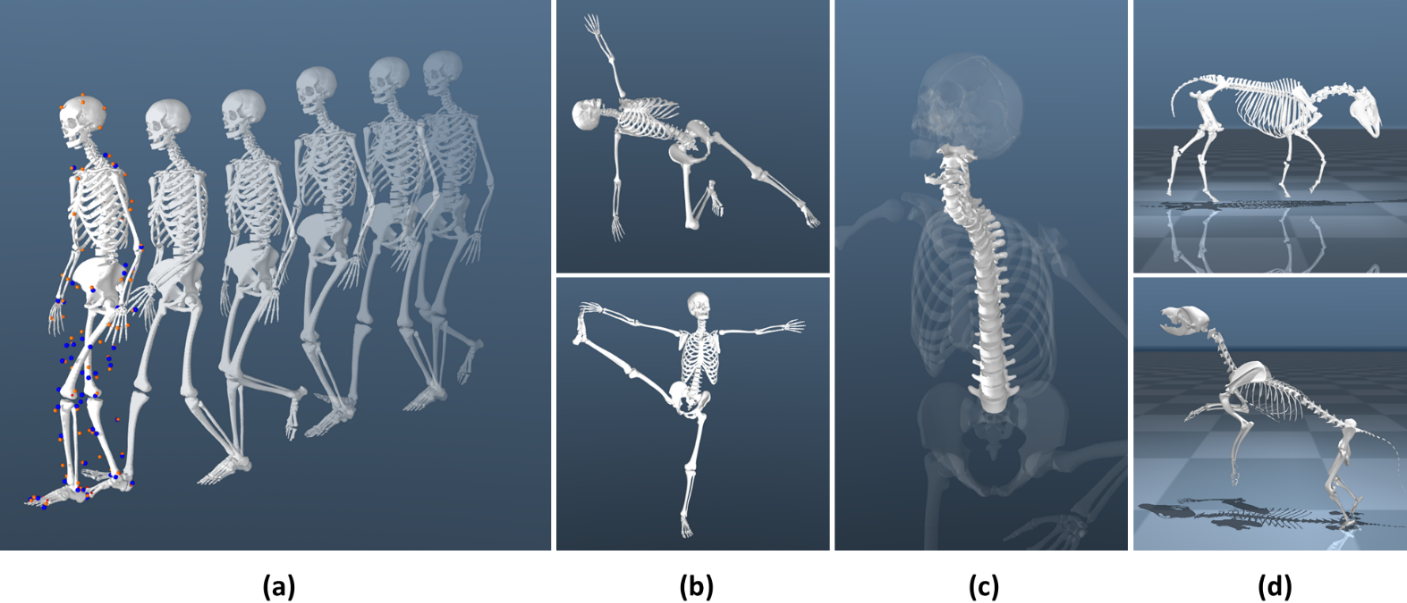}
\caption{Representative outputs of SmoCap motion canonicalization. Motion and scale reconstruction: (a) human gait, (b) yoga poses, (c) proxy-mapped spine motion, and (d) quadruped motion examples. The quadruped examples use public dog motion capture and PFERD horse motion data \cite{counsell2016dogmocap,li2024pferd}.}
\label{fig:word-image-01}
\end{figure*}

Optical marker-based model reconstruction is widely recognized as a constrained optimization problem \cite{lu1999global,andersen2009overdeterminate,duprey2010constraints}. In practical trials, however, it is often mixed-determinate: some DoFs are weakly observed and under-informed, while others are locally over-constrained by dense but inconsistent marker observations. Under-informed parts are not uniquely identifiable, whereas over-constrained parts require the solver to arbitrate among contradictory residuals. This leads to two connected failure modes. First, morphology parameters such as segment dimensions and joint centers enter the same forward kinematics map as pose, and both can explain the same optimization residual. Scaling studies have shown that the recovered morphology affects downstream kinematics, repeatability, and robustness \cite{lund2015scaling,bakke2020shape,bakke2023shape}. When morphology or pose is fixed as a stage-wise precondition, residual can be absorbed by the other along near-singular directions, yielding anatomically invalid residual allocation across sequences, which we refer to as scale--pose leakage. Two-level calibration methods tried to alleviate this effect by wrapping scaling solves around repeated IK solves \cite{charlton2004repeatability,reinbolt2005twolevel,reinbolt2008framework}, while separated scale and pose updates can still leave residual trade-offs. A simultaneous updating strategy incorporates scale update with a Schur complement into pose updates, but requires a good initial guess and substantial variable space and runtime \cite{andersen2010parameter}. Second, coordinated anatomical structures often have more properties than can be inferred from limited observations. Retaining many DoFs preserves expressivity but relies on weight tuning and damping, which can become brittle under marker noise, dropout, or small modeling changes \cite{andersen2009overdeterminate,damsgaard2006anybody}; reducing DoFs improves identifiability at the cost of expressivity for complex joints or diverse shapes \cite{dezee2007lumbar,li2021hybrik,jiang2025manikin}. Whether the problem appears globally over-determinate or locally under-informed, marker fits can remain numerically acceptable without establishing valid morphology or internal movement.

Recent progress has led to more automated and versatile frameworks, yet scale--pose ambiguity, weak observability, and batch-scale reliability remain difficult. OpenCap uses a trained marker reconstruction model followed by scaling and IK in OpenSim \cite{uhlrich2023opencap}. AddBiomechanics formulates scaling and IK as a nonconvex bilevel problem. It relies on a high quality initialization and requires 3--5 minutes for scale and IK \cite{werling2023addbiomechanics}. Differentiable biomechanics enables end-to-end fitting within differentiable simulators but requires several minutes for a typical ten-trial session \cite{cotton2024differentiable}, while MANIKIN uses trained, anatomically constrained, reduced-DoF SMPL representations \cite{loper2015smpl,jiang2025manikin}. These methods shift the cost toward minutes-scale optimization or large-scale training. An inspectable, high-throughput scale--pose canonicalizer that generalizes across articulated models with lightweight configuration and no retraining is still missing.

\begin{figure}[!b]
\centering
\includegraphics[width=\columnwidth]{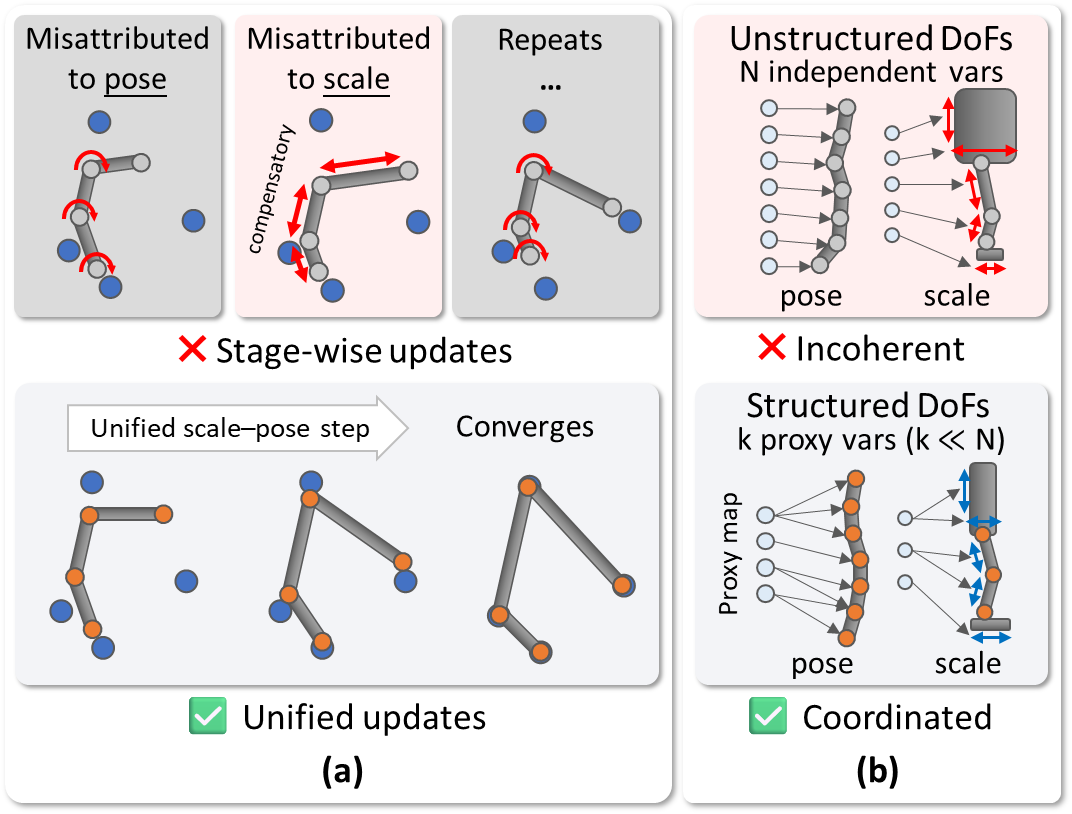}
\caption{Design rationale under mixed determinacy. (a) Stage-wise updates can repeatedly assign residuals to pose or scale, whereas unified local scale--pose updates resolve both components simultaneously. (b) Under weak observability, independent high-DoF controls can produce incoherent internal motion, whereas proxy-coupled low dimensional controls coordinate structured DoFs.}
\label{fig:word-image-02}
\end{figure}

We address this gap with SmoCap, a leakage-resistant unified scale--pose canonicalizer built on explicit trust-region steps. In SmoCap, morphology and posture are resolved in the same local trust-region step, so pose and scale are not forced into sequential precedence; coordinated structures are represented through a low dimensional proxy subspace. Under matched observations, SmoCap is compared with an established OpenSim baseline using fluoroscopy-derived knee motion and anthropometric ground truth as external references rather than marker RMSE alone. Fig. 1 shows representative canonicalization outputs, and Fig. 2 shows the design rationale. The solver runs at a sub-millisecond median per frame and seconds-level for long trials, while an optional light weight root-to-leaf pre-solve provides a coherent warm start for difficult configurations. In summary, the main contributions are (i) a unified scale--pose TR-QP with proxy-mapped coordination under weak observability, (ii) controlled external validation against OpenSim showing that marker RMSE alone does not indicate better movement or morphology recovery, and (iii) training-free cross-model throughput at dataset scale.

\section{Materials and methods}

\subsection{Solver overview}

For each estimation problem, SmoCap requires 3D marker observations $x_{\mathrm{obs}}$ and an articulated model that maps pose generalized coordinates $q$ and morphology scale variables $s$, stacked as $y$, to model marker positions $\hat{x}(y)$. The unknown full state $y$ is represented by low dimensional control variables $p$ with a proxy mapping $y=\Phi(p)$. The joint state keeps pose and scale in the same local step, while the proxy mapping represents coordinated structures. At each iteration, the residuals $r_m(y)$ are linearized to form a box-constrained TR-QP. The resulting step $\Delta p$ is accepted when it gives a positive actual decrease, and the acceptance ratio $\rho$ computed from actual and predicted decreases is used to update the trust-region damping. The estimation terminates when a residual, relative-decrease, or iteration criterion is met and returns $y$.

SmoCap optionally includes (i) full-space soft-reference tracking on $y$ for user-specified priors and (ii) a pre-solve for challenging initial configurations, such as entanglements caused by long-range coupling. The pre-solve provides a TR-QP warm start through root-to-leaf sweeps of overlapping local subproblems with Gauss--Seidel-style updates.

\subsection{Notation}

{\small
\noindent\rule{\columnwidth}{0.4pt}
\noindent\makebox[0.472\columnwidth][l]{\textbf{Symbol}}\textbf{Definition}\par
\noindent\rule{\columnwidth}{0.4pt}
\begin{description}[
  font=\normalfont,
  leftmargin=0.472\columnwidth,
  labelwidth=0.46\columnwidth,
  labelsep=0.012\columnwidth,
  itemsep=1pt,
  parsep=0pt,
  topsep=2pt,
  align=parleft,
  style=multiline,
  before=\raggedright
]
\item[$q\in\mathbb{R}^{n_q}$] Pose generalized coordinates
\item[$s\in\mathbb{R}^{n_s}$] Morphology scale variables
\item[{$y=[q^\top,s^\top]^\top\in\mathbb{R}^{n_y}$}] Full optimization variable
\item[$x_{\mathrm{obs}}\in\mathbb{R}^{n_x}$] Observed marker positions for the current frame
\item[$\hat{x}(y)\in\mathbb{R}^{n_x}$] Model-predicted marker positions
\item[$r_m(y)=\hat{x}(y)-x_{\mathrm{obs}}\in\mathbb{R}^{n_x}$] Marker residual vector
\item[$e_t(y)=y-y_{\mathrm{ref}}\in\mathbb{R}^{n_y}$] Full-state tracking residual
\item[$W_m\in\mathbb{R}^{n_x\times n_x}$] Square-root marker residual weight matrix
\item[$W_t\in\mathbb{R}^{n_y\times n_y}$] Square-root full-space tracking weight matrix
\item[$W_{\mathrm{damp}}\in\mathbb{R}^{n_p\times n_p}$] Proxy-space damping weight matrix
\item[$p\in\mathbb{R}^{n_p}$] Proxy variable in the reduced subspace
\item[$\Phi(p)=y\in\mathbb{R}^{n_y}$] Mapping from proxy space to the full variable space
\item[$J_y=\partial \hat{x}/\partial y\in\mathbb{R}^{n_x\times n_y}$] Marker Jacobian in the full variable space
\item[$J_\Phi=\partial\Phi/\partial p\in\mathbb{R}^{n_y\times n_p}$] Jacobian of the proxy mapping
\item[$J_m=J_yJ_\Phi\in\mathbb{R}^{n_x\times n_p}$] Marker Jacobian in the proxy space
\item[$\Delta p\in\mathbb{R}^{n_p}$] Bounded proxy-space step
\item[$\rho$] Trust-region-style model agreement ratio
\end{description}
\vspace{-0.2em}
\noindent\rule{\columnwidth}{0.4pt}
}

\subsection{TR-QP problem formulation}

SmoCap minimizes the marker objective, optionally with a tracking term, as below

\begin{equation}
\label{eq:word-01}
E(y)=\frac{1}{2}\left\|W_m r_m(y)\right\|_2^2+
\frac{1}{2}\left\|W_t e_t(y)\right\|_2^2 .
\end{equation}

The problem is defined in the full state $y$, and optimized with proxy variable $p$ through the mapping $\Phi$, i.e., $\tilde{E}(p):=E(\Phi(p))$. The proxy map $\Phi$ specifies how reduced subspace variables control full pose--scale coordinates. Since $y=[q^\top,s^\top]^\top$, the full marker Jacobian can be written as $J_y=[J_q\ J_s]$, where $J_q$ is the analytical pose Jacobian and $J_s$ is the analytical scale Jacobian. Thus, before proxy projection, the local marker residual increment is $\Delta r_m \approx J_q\Delta q+J_s\Delta s$, which ensures that morphology enters the same Cartesian residual space as posture.

At each iterate, the marker residual is linearized as

\begin{equation}
\label{eq:word-02}
r_m(y+\Delta y)\approx r_m(y)+J_y\Delta y,\qquad
\Delta y\approx J_\Phi\Delta p .
\end{equation}

which yields the proxy-space approximation

\begin{equation}
\label{eq:word-03}
r_m\!\left(\Phi(p+\Delta p)\right)
\approx r_m\!\left(\Phi(p)\right)+J_m\Delta p .
\end{equation}

where $J_m=J_yJ_\Phi$. For $e_t(y)=y-y_{\mathrm{ref}}$, the tracking increment is $J_\Phi\Delta p$.

SmoCap constructs the quadratic model in $\Delta p$ as

\begin{equation}
\label{eq:word-04}
\begin{aligned}
m(\Delta p)=
&\frac{1}{2}\left\|W_m\left(J_m\Delta p+r_m\right)\right\|_2^2 \\
&+\frac{1}{2}\left\|W_t\left(J_\Phi\Delta p+e_t\right)\right\|_2^2 \\
&+\frac{1}{2}\left\|W_{\mathrm{damp}}^{1/2}\Delta p\right\|_2^2 .
\end{aligned}
\end{equation}

where $r_m$, $e_t$, $J_y$, $J_\Phi$, and $J_m$ are evaluated at the current iterate, $W_m$ and $W_t$ are the marker and tracking square-root weight matrices, and $W_{\mathrm{damp}}$ applies per-coordinate regularization in the proxy space.

Expanding $m(\Delta p)$ gives the standard QP form

\begin{equation}
\label{eq:word-05}
\min_{\Delta p}\ \frac{1}{2}\Delta p^\top H\Delta p+g^\top\Delta p .
\end{equation}

where

\begin{equation}
\label{eq:word-06}
\begin{aligned}
H&=J_m^\top W_m^\top W_m J_m
+J_\Phi^\top W_t^\top W_t J_\Phi+W_{\mathrm{damp}},\\
g&=J_m^\top W_m^\top W_m r_m+J_\Phi^\top W_t^\top W_t e_t .
\end{aligned}
\end{equation}

The trust-region step solves a proxy-space QP with proxy-step bounds

\begin{equation}
\label{eq:word-07}
\begin{alignedat}{2}
\Delta p^\star
&=\arg\min_{\Delta p}\quad &&\frac{1}{2}\Delta p^\top H\Delta p+g^\top\Delta p,\\
&\text{subject to}\quad &&\ell_b\leq \Delta p\leq u_b .
\end{alignedat}
\end{equation}

Here $\ell_b$ and $u_b$ are lower and upper proxy-step bounds. The predicted decrease is

\begin{equation}
\label{eq:word-08}
\Delta f_{\mathrm{pred}}
=-g^\top\Delta p^\star-\frac{1}{2}(\Delta p^\star)^\top (H-W_{\mathrm{damp}})\Delta p^\star .
\end{equation}

The actual decrease used for step acceptance is

\begin{equation}
\label{eq:word-09}
\Delta f_{\mathrm{act}}=
\tilde{E}(p)-\tilde{E}(p+\Delta p^\star).
\end{equation}

The acceptance ratio is computed as

\begin{equation}
\label{eq:word-10}
\rho=\frac{\Delta f_{\mathrm{act}}}{\Delta f_{\mathrm{pred}}+\epsilon_\rho}.
\end{equation}

where $\epsilon_\rho>0$ is a small numerical stabilizer. If the actual decrease is non-positive, the trial step is rejected and shrunk; otherwise the step is accepted with $p\leftarrow p+\Delta p^\star$, followed by $y\leftarrow\Phi(p)$. The ratio $\rho$ is used to update the damping.

\subsection{Implementation and complexity}

Each iteration evaluates $\hat{x}(y)$ and residuals for $m$ markers, computes Jacobians to form $J_m$, assembles $(H,g)$, solves a bounded proxy-space QP in $n_p$ variables, and evaluates the step acceptance decrease. The dominant operations scale with marker count through forward kinematics and Jacobian evaluation, and with proxy dimension through the TR-QP size. The implementation uses MuJoCo for articulated model representation, kinematics, and box-constrained QP routines \cite{todorov2012mujoco}; additional implementation details are provided in the project repository, including JIT compilation, logging schema, engineering optimizations, and dataset-specific preprocessing.

\subsection{Datasets and reference measurements}

We use three datasets for external movement and morphology validation and coordinated-structure evaluation under weak observability. The CAMS-Knee evaluation includes six subjects and 28 movement trials with synchronized optical marker trajectories, fluoroscopy-derived femur and tibia motion measurements, and anthropometric measurements \cite{taylor2017cams}. The Riglet dataset provides optical marker trajectories together with session anthropometric measurements from 30 subjects \cite{riglet2024dataset}. Fifty-seven yoga trials with extreme-pose marker trajectories from the MoYo dataset were used for the coordinated-spine ablation \cite{tripathi2023ipman}. Dataset-specific preprocessing, synchronization, alignment, and trial-selection procedures are described in the project repository.

\begin{figure}[!t]
\centering
\includegraphics[width=0.702\columnwidth]{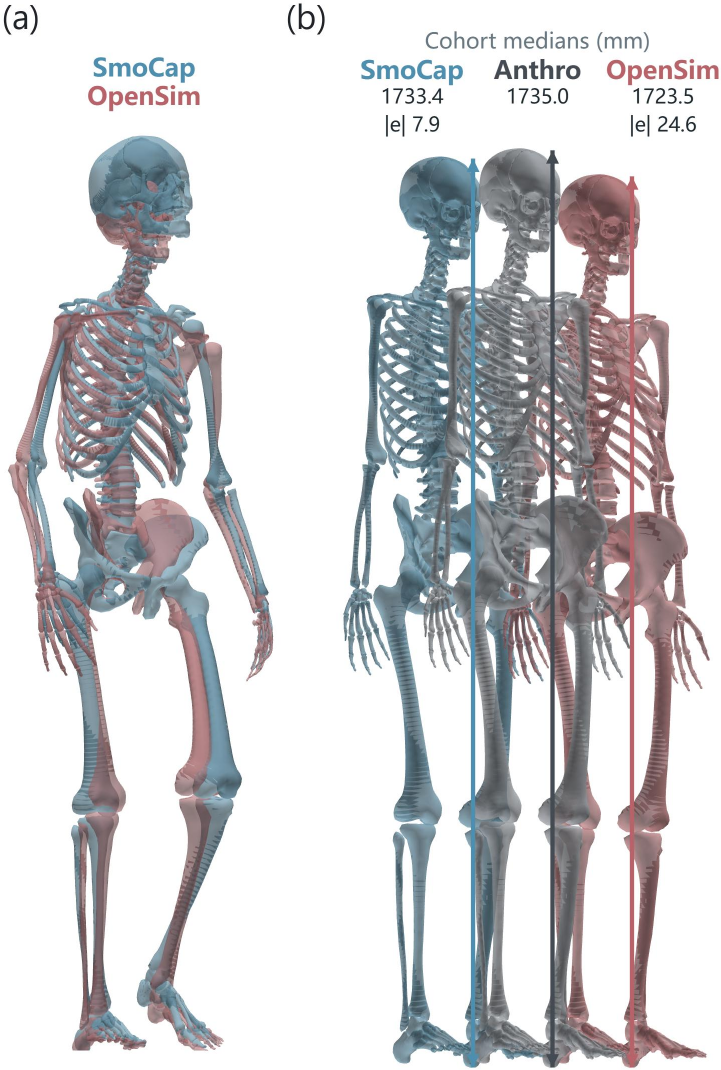}
\caption{Visual comparison of SmoCap and OpenSim reconstructions. (a) representative overlaid reconstruction. (b) Cohort-median height comparison. Displayed deviations are magnified $2\times$.}
\label{fig:smocap-opensim-visual}
\end{figure}

\begin{figure*}[!b]
\centering
\captionsetup{skip=6pt}
\includegraphics[width=0.9\linewidth]{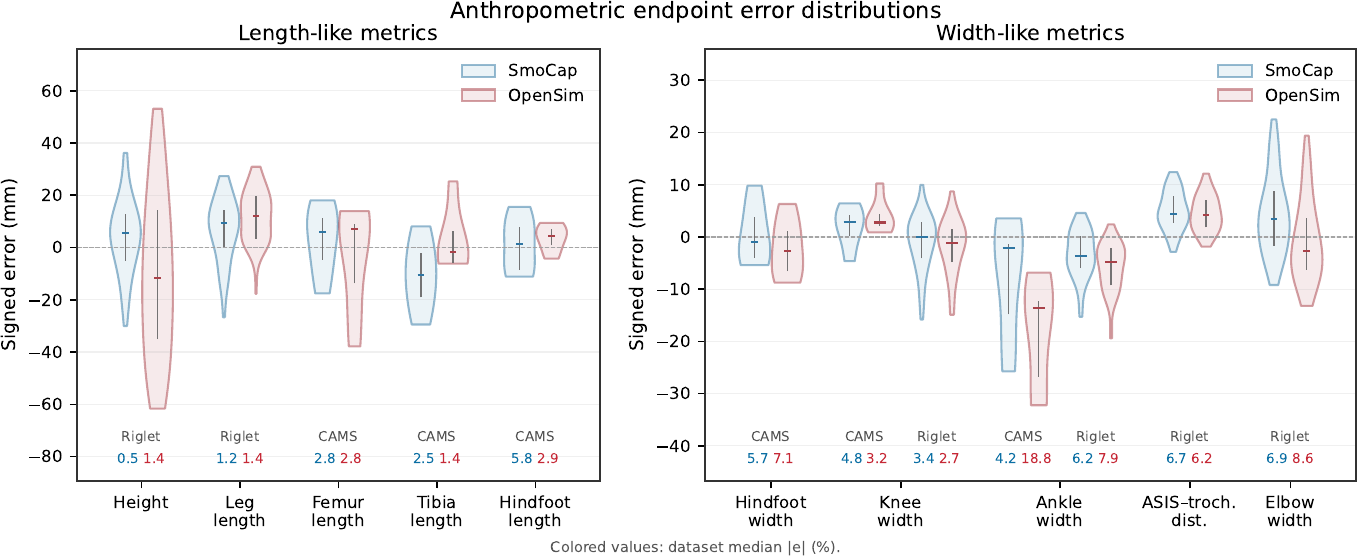}
\begingroup
\renewcommand{\thefigure}{5}
\caption{Anthropometric endpoint error distributions for external morphology validation. Paired SmoCap and OpenSim signed endpoint errors are shown for length-like and width-like metrics on CAMS-Knee and Riglet datasets. Violin distributions summarize endpoint errors, and the colored values report dataset median absolute percentage errors.}
\label{fig:word-image-04}
\endgroup
\end{figure*}

\begin{table}[!t]
\centering
\includegraphics[width=\columnwidth]{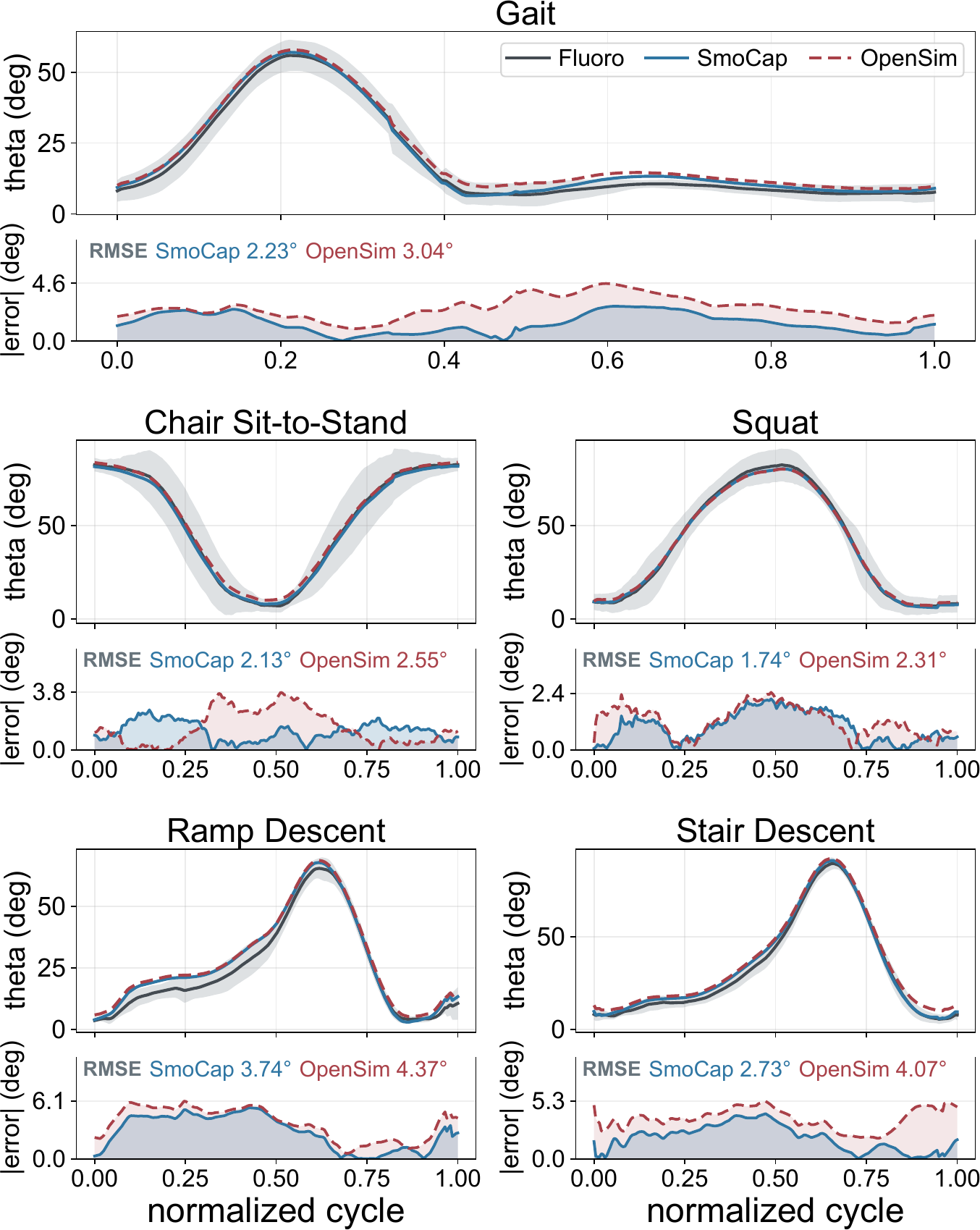}
\begingroup
\renewcommand{\thefigure}{4}
\captionof{figure}{Fluoroscopy-based knee flexion comparison. SmoCap and OpenSim trajectories are shown with fluoroscopy-derived references over normalized cycles across five movements. Shaded regions show cross-trial variation; the lower panels show absolute errors and activity-level curve RMSE.}
\label{fig:word-image-03}
\endgroup
\end{table}

\subsection{Evaluation and reporting}

The evaluation is threefold: external movement and morphology validity, coordinated structure under weak observability, and throughput. For external validity, we report marker fit together with external pose and morphology references \cite{hicks2015model}. Pose is evaluated as Grood--Suntay knee orientation derived from fluoroscopy \cite{grood1983joint}, and morphology is evaluated against direct anthropometric endpoints. SmoCap was compared with the OpenSim ScaleTool--MarkerPlacer--IK workflow \cite{delp2007opensim,kainz2017scaling} using the same articulated musculoskeletal model implemented in MuJoCo and OpenSim. Both pipelines received identical static and dynamic marker observations with unit weights. OpenSim adjusted marker placement with MarkerPlacer as part of its standard workflow, while SmoCap markers remained unchanged \cite{dunne2021marker,price2020marker}. Each pipeline estimated subject scale from the static observations and then performed IK with scale fixed. The three knee DoFs were unclamped in both pipelines for full motion recovery assessment. Morphology was evaluated from homologous reconstructed model quantities in a neutral pose. Repeated trials were averaged within subject for the primary comparison.

For coordinated structure preservation, we use MoYo dataset to perform a spine DoFs coupling ablation \cite{dezee2007lumbar}. Three configurations were compared: proxy variables controlled DoFs full spine via polynomial mapping (poly), independent DoFs full spine (nopoly), and a reduced three-segment spine (classical). We report marker error, coordination plausibility statistics such as spatial and temporal roughness of spine DoFs, and representative bending-pattern visualizations.

For throughput and convergence statistics, we report per- frame and dataset level runtime, together with solver statistics.

\section{Results}

We report external validity, proxy driven behavior under weak observability, and throughput. A direct visual comparison of SmoCap and OpenSim is shown in Fig. 3.

\vspace{-4pt}
\subsection{Fluoroscopy and anthropometry-based external validation}

Aggregated fluoroscopy kinematics references and predicted knee flexion trajectories are shown in Fig. 4. SmoCap achieved lower activity-level curve RMSE than OpenSim across all five movements. In the controlled comparison, the OpenSim baseline achieved lower marker RMSE (8.81 vs 19.14 mm), whereas SmoCap achieved lower knee-orientation RMSE against fluoroscopy (5.82$^\circ$ vs 7.50$^\circ$) in 24 of 28 trials and all six subjects.

\begin{table}[!ht]
\captionsetup{skip=2pt}
\caption{Contextual comparison with reported tibiofemoral flexion validation.}
\label{tab:word-13}
\centering
\small
\setlength{\tabcolsep}{3pt}
\renewcommand{\arraystretch}{1.05}
\begin{tabular*}{\columnwidth}{@{\extracolsep{\fill}}lccr@{}}
\hline
Research & \multicolumn{1}{c}{Non-invasive} & \multicolumn{1}{c}{Deep flexion} & \multicolumn{1}{c}{FE error ($^\circ$)} \\
\hline
SmoCap & Yes & Yes & 2.9 \\
Reinschmidt et al. \cite{reinschmidt1997skeletal} & No & No & 2.1 \\
Tranberg et al. \cite{tranberg2011rsa} & No & No & 2--5 \\
F\"andriks et al. \cite{fandriks2025skin} & No & Yes & 5--15 \\
Wang et al. \cite{wang2021portable} & Yes & No & 1.4 \\
\hline
\end{tabular*}
\end{table}

Anthropometric endpoint error distributions are shown in Fig. 5 for CAMS-Knee and Riglet datasets. Although individual metrics varied, mean absolute anthropometric endpoint errors were lower for SmoCap on both datasets: 8.44 vs 9.51 mm on CAMS-Knee and 6.87 vs 8.77 mm on Riglet.

Table II summarizes the matched movement and morphology results.

As an independent contextual comparison with prior ground-truth studies, Table I compares the SmoCap FE result against reported validation studies. Typical workflows for obtaining skeletal ground truth rely on invasive measures such as bone pins or implanted tantalum markers \cite{reinschmidt1997skeletal,tranberg2011rsa,fandriks2025skin}. In contrast, our evaluation is performed only with skin markers, while covering deep-flexion functional activities including squat and sit to stand. The reported FE error in our results (RMSE $\approx$ 2.9$^\circ$) is on the same level as classical walking validations (e.g., approximately 2$^\circ$ mean difference in bone-pin studies), and is substantially lower than those reported in high flexion RSA validations (5--15$^\circ$ mean differences).

Marker RMSE alone did not reliably indicate better movement or morphology recovery. Despite its higher marker RMSE, SmoCap achieved lower knee-orientation and aggregate anthropometric errors in the matched comparison.

\begin{table}[!t]
\caption{External movement and morphology validation against OpenSim.}
\label{tab:word-14}
\centering
\small
\setlength{\tabcolsep}{1.5pt}
\renewcommand{\arraystretch}{1.18}
\begin{tabular*}{\columnwidth}{@{\extracolsep{\fill}}lccccc@{}}
\hline
 & \multicolumn{4}{c}{CAMS-Knee \cite{taylor2017cams}} & Riglet \cite{riglet2024dataset} \\
\cline{2-5}\cline{6-6}
Method & Marker & Orient. & Trial & Anthro. & Anthro. \\
 & RMSE & RMSE & wins & MAE & MAE \\
 & (mm) & ($^\circ$) & & (mm) & (mm) \\
\hline
SmoCap & 19.14 & \textbf{5.82} & \textbf{24/28} & \textbf{8.44} & \textbf{6.87} \\
OpenSim & \textbf{8.81} & 7.50 & 4/28 & 9.51 & 8.77 \\
\hline
\end{tabular*}
\end{table}

\begin{figure}[!t]
\centering
\captionsetup{skip=18pt}
\includegraphics[width=\columnwidth]{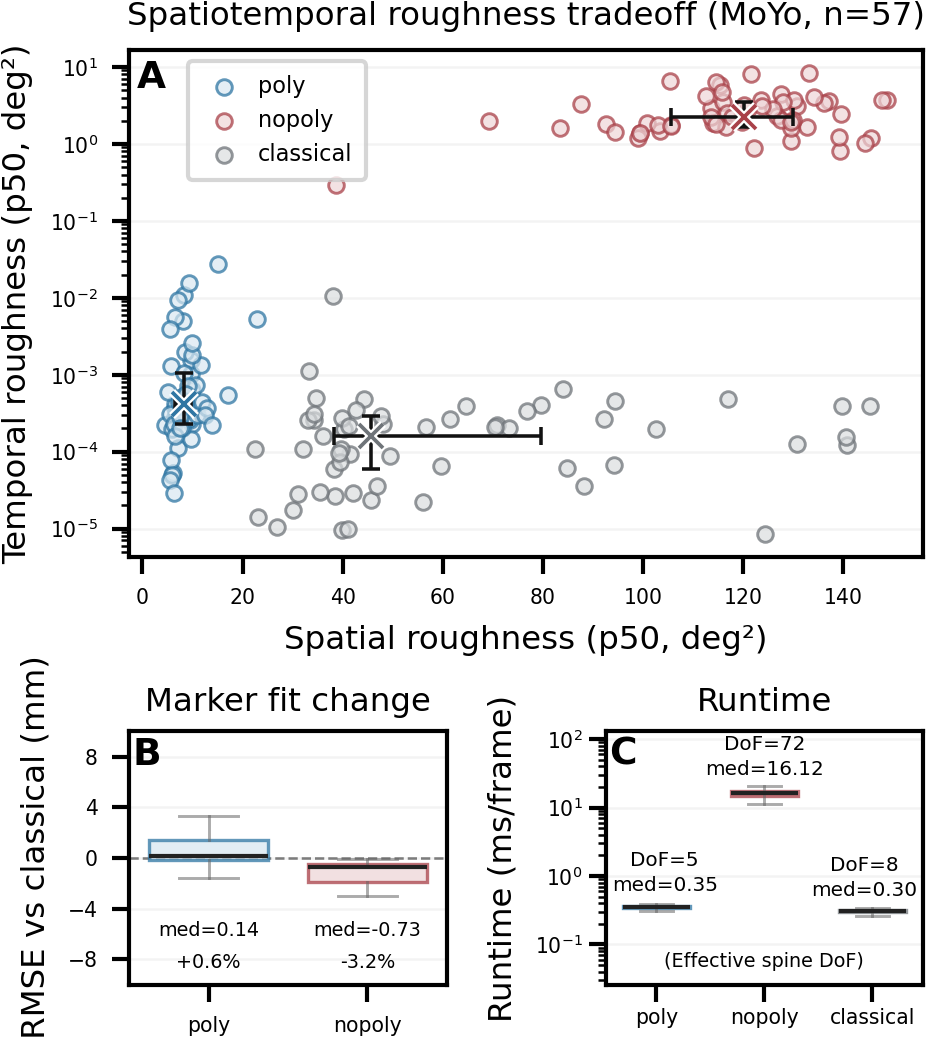}
\caption{Proxy ablation under weak observability. (A) Spatiotemporal spine roughness for the proxy-coupled spine (poly), independent-DoF spine (nopoly), and segmented spine (classical). (B) Marker-fit change relative to the segmented-spine baseline. (C) Runtime across the three spine configurations.}
\label{fig:word-image-05}
\vspace{10pt}
\includegraphics[width=\columnwidth]{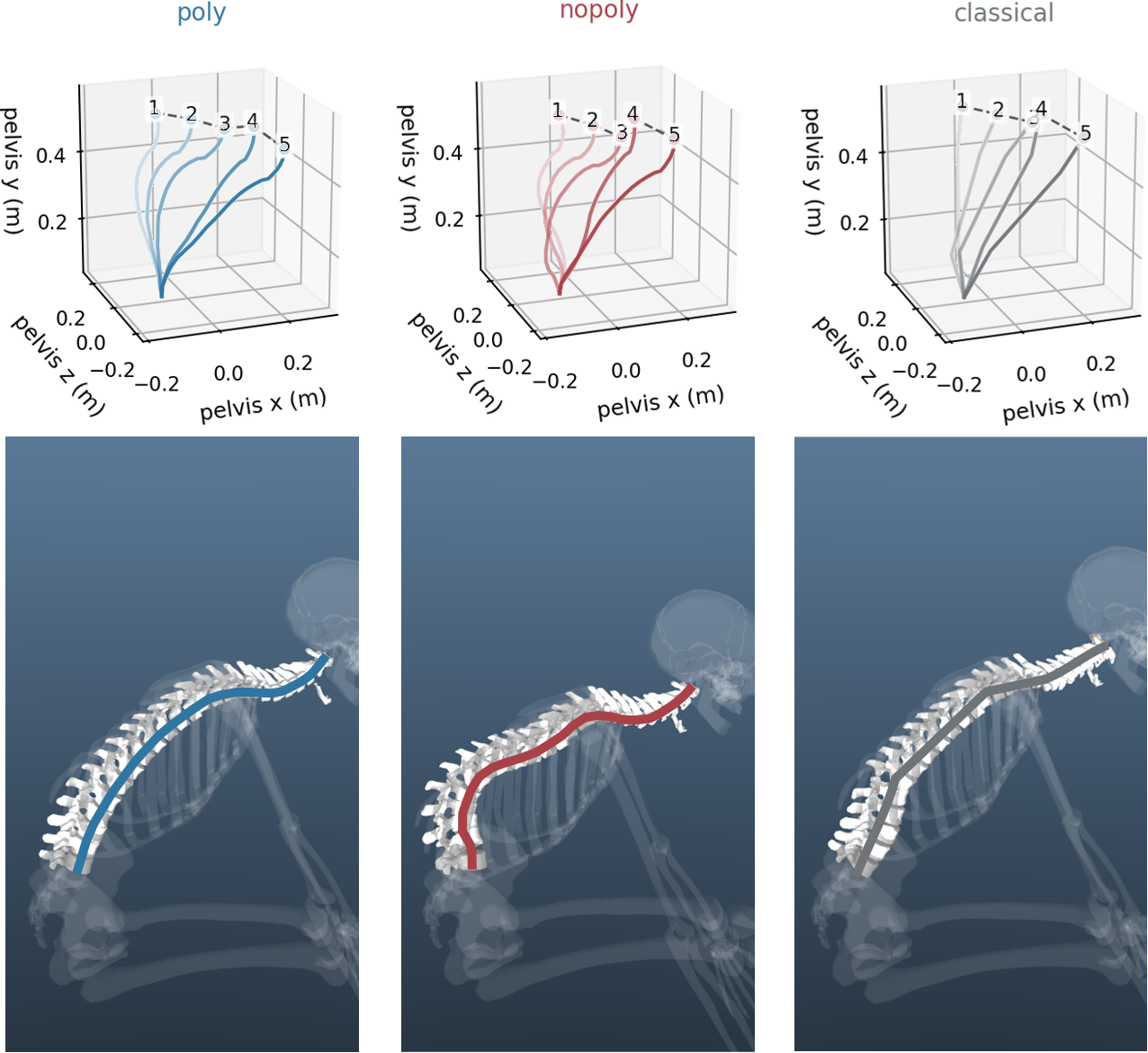}
\caption{Spine bending patterns under different spine configurations. Top: spine curves at five sampled time points. Bottom: skeleton render at a representative frame. The proxy-coupled spine (poly) shows smooth distributed curvature, the independent-DoF spine (nopoly) shows fragmented local bending, and the segmented spine (classical) shows piecewise rigid bending.}
\label{fig:word-image-06}
\end{figure}

\subsection{Proxy ablation under weak observability for naturally coordinated structures}

The MoYo dataset includes extreme and large-amplitude poses but only sparse surface markers, leaving the multi-DoF yet naturally coordinated spine indirectly constrained \cite{tripathi2023ipman}. We compared proxy-mapped full-spine DoFs (poly), independent full-spine DoFs (nopoly), and a reduced three-segment spine (classical) on the same sequences. Because MoYo has no ground-truth spine reference, roughness and bending-pattern measures are treated as solver-behavior proxies rather than biological ground truth. Marker fit, spatiotemporal roughness, and runtime are summarized in Fig. 6.

Fig. 6A shows the temporal roughness of per joint angle history, and spatial roughness of spine at each time instant. Poly exhibits lower roughness overall. In contrast, nopoly formed a separated high-roughness cluster, whereas classical was temporally smooth but spatially rougher.

Fig. 6B reports relative marker fitting error of poly and nopoly against classical. Poly is close to classical (median $\Delta$RMSE = +0.14 mm, +0.6\%), whereas nopoly shows a mild reduction in RMSE (median $\Delta$RMSE = -0.73 mm, -3.2\%). Thus, lower roughness can be achieved with configurations with similar marker errors.

Fig. 6C compares runtime across configurations. Median runtime was 0.35 ms/frame for poly (DoFs = 5), 0.30 ms/frame for classical (DoFs = 8), and 16.12 ms/frame for nopoly (DoFs = 72). Poly and classical remain at a similar level, while nopoly leads to substantially higher computational cost.

Fig. 7 shows distinct spine bending patterns: poly is smooth and globally distributed, nopoly shows fragmented and poorly coordinated local bending, and classical exhibits a piecewise and broken pattern with rigid segments.

Taken together, Figs. 6--7 show that under weak observability, poly, nopoly, and classical can achieve similar marker fits while yielding distinct spine patterns. Proxy coupling retains expressive curvatures and leads to more coordinated behavior without materially increasing fitting error or runtime relative to classical.

\subsection{Throughput and convergence across datasets}

Table III summarizes throughput across CAMS-Knee, Riglet, and MoYo datasets. Median runtime was 0.204--0.332 ms/frame, with p90 below 0.5 ms and consistently 2--3 iterations across datasets. For reference, a 10-s, 100-Hz trial requires about 0.2--0.3 s of pure solver time.

\begin{table}[!t]
\caption{Default throughput and convergence statistics across datasets.}
\label{tab:word-15}
\centering
\small
\setlength{\tabcolsep}{2pt}
\renewcommand{\arraystretch}{1.16}
\begin{tabular*}{\columnwidth}{@{\extracolsep{\fill}}lrrrrrrrr@{}}
\hline
\multirow{2}{*}{Dataset} & \multirow{2}{*}{Frames} & \multirow{2}{*}{FPS} & \multicolumn{2}{c}{Time (ms)} & \multicolumn{2}{c}{Iters} & \multicolumn{2}{c}{RMSE (mm)} \\
\cline{4-9}
 &  &  & \multicolumn{1}{c}{p50} & \multicolumn{1}{c}{p90} & \multicolumn{1}{c}{p50} & \multicolumn{1}{c}{p90} & \multicolumn{1}{c}{p50} & \multicolumn{1}{c}{p90} \\
\hline
CAMS-Knee & 10740 & 4902 & 0.204 & 0.301 & 2 & 3 & 18.2 & 20.6 \\
Riglet & 10138 & 3353 & 0.298 & 0.383 & 3 & 3 & 17.8 & 21.1 \\
MoYo & 11698 & 3013 & 0.332 & 0.498 & 2 & 3 & 25.8 & 32.4 \\
\hline
\end{tabular*}
\vspace{2pt}
\begin{minipage}{\columnwidth}
\footnotesize Per-frame solver runtime with file I/O and model loading excluded.
\end{minipage}
\end{table}

Table IV summarizes the optional pre-solve warm start. Pre-solve increased cold-start runtime by around 3--7 times across CAMS-Knee, Riglet, and MoYo, but consistently reduced outlier rates. MoYo showed the largest reduction, from 18.33\% to 5.93\%, while CAMS-Knee and Riglet decreased to 1.87\% and 0\%, respectively. Overall, SmoCap performed stably in regular cases, while the optional pre-solve reduced failures in difficult configurations.

\begin{table}[!t]
\caption{Pre-solve warm-start performance.}
\label{tab:word-16}
\centering
\small
\setlength{\tabcolsep}{1pt}
\renewcommand{\arraystretch}{1.16}
\begin{tabular*}{\columnwidth}{@{\extracolsep{\fill}}lrrrrrrr@{}}
\hline
\multirow{2}{*}{Dataset} & \multirow{2}{*}{Frames} & \multicolumn{1}{c}{Overhead} & \multicolumn{2}{c}{Pre (ms)} & \multicolumn{3}{c}{Out. (\%)} \\
\cline{4-8}
 &  & \multicolumn{1}{c}{($\times$)} & \multicolumn{1}{c}{p50} & \multicolumn{1}{c}{p90} & \multicolumn{1}{c}{No-pre} & \multicolumn{1}{c}{Pre} & \multicolumn{1}{c}{$\Delta$} \\
\hline
CAMS-Knee & 1500 & 3.28 & 5.010 & 6.221 & 4.53 & 1.87 & 2.67 \\
Riglet & 1500 & 4.54 & 9.127 & 10.288 & 0.13 & 0 & 0.13 \\
MoYo & 1500 & 7.39 & 12.677 & 14.527 & 18.33 & 5.93 & 12.40 \\
\hline
\end{tabular*}
\vspace{2pt}
\begin{minipage}{\columnwidth}
\footnotesize Outliers exceed the no-pre median by 6 MADs within each dataset; $\Delta$ is the reduction in outlier rate.
\end{minipage}
\end{table}

\FloatBarrier
\section{Discussion}

Marker RMSE alone did not reliably indicate better motion or morphology recovery \cite{hicks2015model}. In the controlled comparison, the OpenSim baseline achieved lower marker RMSE, whereas SmoCap achieved lower knee-orientation RMSE against fluoroscopy and lower mean absolute anthropometric endpoint errors on both datasets. This exposes the core ambiguity of marker-based reconstruction: the same fitting error can be explained by different mixtures of morphology and posture, while weakly observed DoFs are not uniquely identifiable. SmoCap addresses the two parts of this problem by resolving morphology and posture in the same local trust-region step and controlling coordinated structures through a low dimensional proxy subspace. The external evaluations test reconstructed movement and morphology beyond marker fit; the spine ablation probes coordinated behavior under weak observability; and the throughput evaluation tests whether this control remains practical at dataset scale.

Fluoroscopy and anthropometric data provide independent external measurements of kinematics and scale \cite{benoit2006skin,miranda2013biplanar}. The calculated knee kinematics align closely with fluoroscopy across tasks including deep-flexion activities, while anthropometric endpoint errors remain small on CAMS-Knee and Riglet. These measurements also allow us to check marker fit, movement validity, and morphology validity separately. The comparison used the same musculoskeletal model and identical static and marker observations with unit weights. OpenSim adjusted marker placement with MarkerPlacer as part of its standard workflow, whereas SmoCap markers remained unchanged \cite{dunne2021marker,price2020marker}. Under this setting, the lower marker residual of the OpenSim workflow did not indicate lower external knee-orientation or anthropometric errors. SmoCap retained a fixed dataset-level marker definition and did not perform subject-specific manual marker adjustment or marker-offset optimization because manual adjustment harms scalability to large datasets, while marker-offset optimization is highly under-constrained and can absorb residual without identifying sources. Taken together, the external results support the finding that SmoCap achieved lower movement and morphology errors with its unified scale--pose formulation, despite its higher marker residual.

The spine ablation under extreme yoga poses shows why proxy coupling is important under weak observability. The spine has many DoFs that are not fully independent, yet ordinary sparse marker sets cannot fully constrain them \cite{dezee2007lumbar}. This creates a tradeoff between the less expressive classical model with segmented rigid spines and the less predictable and computationally heavy nopoly model with full DoFs. The poly model instead drives the free DoFs through proxy variables while retaining expressive curvatures. Figs. 6--7 show that all three models can achieve similar marker fits while yielding distinct spine patterns in roughness, morphology, and computation. Compared with classical, poly changes marker RMSE only marginally by +0.14 mm (+0.6\%), but produces lower roughness at near-identical runtime (0.35 vs 0.30 ms/frame). Nopoly achieves the lowest marker fitting error but produces fragmented local bending patterns and requires roughly 50-fold higher runtime (16.12 ms/frame). Thus, expressiveness cannot be achieved simply by adding DoFs; weakly observed internal freedom also needs to be coordinated. Proxy coupling provides this control while keeping fitting error and runtime close to the classical baseline.

Throughput is a key feature of SmoCap because dataset-scale canonicalization must be cheap enough to become routine. Computational speed remains a practical limitation in model fitting pipelines \cite{charlton2004repeatability,reinbolt2005twolevel,andersen2010parameter,hammond2025pipeline}. SmoCap runs at sub-millisecond per frame with only 2--3 iterations, so a typical 10 s, 100 Hz trial requires about 0.2--0.3 s of pure solver time. This makes batch processing practical as a standard preprocessing primitive rather than a costly offline calibration step.

A remaining practical bottleneck is extreme configuration reconstruction. Most trials converge under the default solve, while twisted configurations or weak initialization can still lead to poor local minima. The optional pre-solve is designed for these cases. Its Gauss--Seidel-style root-to-leaf local sweeps can untangle complex configurations, especially when constraints are present. Runtime grows several-fold from an already very low base, making it applicable when difficult configurations or suspicious residuals are observed. On MoYo, it reduced the outlier rate from 18.33\% to 5.93\%. It therefore provides a selective recovery path without changing the default high-throughput solve.

Three limitations remain. First, the pipeline relies on corresponding marker definitions and a rigid marker-offset assumption. Marker placement error and soft tissue artifact can therefore reduce robustness \cite{leardini2005sta,benoit2006skin,barre2013sta,andersen2012sta}. Second, inertia-related and musculotendinous scaling were outside the scope of morphology--motion canonicalization. This limits direct use for dynamic analysis, although body-mass scaling and actuator parameter tuning can be incorporated using existing methods \cite{dembia2020moco,sturdy2022rra,hammond2025pipeline,fregly2008cost,heinen2016mtu,modenese2016musculotendon}. Third, the spine ablation evaluates plausibility through spatiotemporal roughness and morphology rather than imaging-based or statistical ground truth. These metrics are informative for solver behavior, but they do not certify biological correctness.

Future work should reduce the rigid marker-transform assumption with richer anatomy-aware priors or models, such that scale and pose remain identifiable when marker placement is uncertain. Dynamics-related attributes should also be included under the same unified formulation to extend SmoCap beyond morphology--motion canonicalization without breaking kinematic--dynamic consistency.

Overall, SmoCap provides a unified and scalable formulation for motion canonicalization that resolves morphology and motion within the same optimization framework and coordinates weakly observed DoFs through proxy coupling. The external evaluations showed that marker RMSE alone did not reliably indicate better movement or morphology recovery, while SmoCap achieved lower errors in both. With sub-millisecond throughput and an optional light weight pre-solve for extreme configurations, SmoCap supports weak-observability-aware, cohort-scale movement reconstruction for downstream pipelines that depend on interpretable subject motion and morphology.

\section*{Data and Code Availability}

The reproducibility package, scripts for reproducing the reported figures and tables, and result artifacts are available at: \url{https://github.com/lshdlut/smocap-paper-artifact}. The Python implementation of the SmoCap solver is available at: \url{https://github.com/lshdlut/smocap-paper-code}. An interactive demonstration is available at: \url{https://lshdlut.com/en/projects/smocap/}. The original datasets are not redistributed; dataset access follows the licenses and access policies of the original data providers.

\section*{Competing interests}

None.

\end{document}